%% file: main.tex
\documentclass[conference]{IEEEtran}
\IEEEoverridecommandlockouts
\makeatletter
\def\endthebibliography{%
  \def\@noitemerr{\@latex@warning{Empty `thebibliography' environment}}%
  \endlist
}
\makeatother

\ifCLASSINFOpdf
	\usepackage[pdftex]{graphicx}
	\graphicspath{{./Images/}}
\else
	\usepackage[dvips]{graphicx}
	\graphicspath{{./Images/}}
\fi
\usepackage{amsmath}
\usepackage{amssymb}
\usepackage{amsfonts}
\usepackage{wasysym}
\usepackage{mathrsfs}
\usepackage{latexsym}
\usepackage{xfrac}
\usepackage{xcolor}
\usepackage{tikz}
\usepackage{booktabs}
\usepackage{multirow}

\usepackage{array}

\usepackage{xr}

\definecolor{myMATLABblue}{RGB}{0, 110, 191}
\definecolor{myFlashRed}{RGB}{255, 0, 0}

\begin{document}
\setlength{\abovedisplayskip}{4pt plus 2pt minus 0pt}
\setlength{\belowdisplayskip}{4pt plus 2pt minus 0pt}
\setlength{\abovecaptionskip}{4pt plus 2pt minus 0pt}
\setlength{\belowcaptionskip}{4pt plus 2pt minus 0pt}
\setlength{\textfloatsep}{8pt plus 2pt minus 0pt}
\setlength{\floatsep}{4pt plus 1pt minus 0pt}
\setlength{\dbltextfloatsep}{8pt plus 2pt minus 0pt}
\setlength{\dblfloatsep}{4pt plus 1pt minus 0pt}
\input{paper_structure.tex}

\end{document}

%% file: paper_structure.tex
\title{A 25-$\mu$s/inf Event-driven Graph Neural Network Processor with Spatiotemporal Caching and Spline Convolution for Ultra-low-latency AI at the Edge

\thanks{This work was funded in part by the Dutch government and by Prophesee as an HTSM-TKI project. Circuit fabrication was supported by TSMC via the University Shuttle Program. Authors with $^*$ share equal credits on the work. Corresponding author: A.~Kneip (a.kneip@tudelft.nl).}
}
\vspace{-0.5cm}

\author{\IEEEauthorblockN{Adrian Kneip$^{1,2}$,~\IEEEmembership{Member, IEEE}, Martin Lefebvre$^{1}$,~\IEEEmembership{Member, IEEE}, Daniel Gehrig$^{3,4}$,~\IEEEmembership{Member, IEEE},\\
Victoria Catalán Pastor$^{3}$,~\IEEEmembership{Graduate Student Member, IEEE}, Davide Scaramuzza$^{3}$,~\IEEEmembership{Senior Member, IEEE},\\ Marian Verhelst$^{2,*}$,~\IEEEmembership{Fellow, IEEE}, and Charlotte Frenkel$^{1,*}$,~\IEEEmembership{Member, IEEE}}\\

\IEEEauthorblockA{$^1$Delft University of Technology (TU Delft), 2628 CD Delft, The Netherlands.
$^2$KU Leuven, 3000 Leuven, Belgium.\\
$^3$University of Zürich (UZH), 8050 Zürich, Switzerland.
$^4$University of Pennsylvania, Philadelphia, PA 19104, USA.}
\vspace{-0.8cm}
}

\maketitle

\input{Sections/abstract}

\input{Sections/intro}
\input{Sections/datapath}
\input{Sections/mem}
\input{Sections/meas}

\input{Sections/conclusion}

%\section*{Acknowledgment}

\bibliographystyle{IEEEtran}
\bibliography{./biblio}

%% file: Sections/abstract.tex
\begin{abstract}
%Event-driven graph neural networks (EV-GNNs) can efficiently leverage the micro-second level resolution of dynamic-vision-sensor (DVS) cameras to provide ultra-low-latency updates in edge-AI detection tasks. However, they require dedicated HW processors capable of simultaneously leveraging their inherent mix of dense-regular compute operations and sparse-irregular memory accesses. To that end, we present ETHEREAL, the first EV-GNN processor chip, scalable to high-resolution tasks (i.e., DVS with up to 1280$\times$720 pixels). It addresses the aforementioned challenges by concurrently introducing a neighbor-parallel spline-convolution engine as well as a 2D/3D-split memory hierarchy with a novel region-of-interest spatiotemporal caching mechanism. Measurement results showcase an end-to-end latency (resp. energy) per event inference of 10$\mu$s (resp. 1$\mu$J) on state-of-the-art workloads such as DAGr-GNN, a 100$\times$ improvement over prior system-level works. 
Dynamic-vision-sensor (DVS) cameras generate events on a per-pixel basis with a $\mu$s-level temporal resolution, calling for new algorithm-hardware co-design approaches compared to standard frame-based vision. While event-driven graph neural networks (EV-GNNs) emerge as a promising algorithmic solution, %algorithm, 
they raise new HW challenges by mixing dense-regular compute operations and sparse-irregular memory accesses. %, thereby lacking hardware support. 
We present ETHEREAL, the first EV-GNN accelerator that scales to 640$\times$480 resolutions, thanks to a neighbor-parallel spline-convolution engine and a 2D/3D-split memory hierarchy with a novel region-of-interest spatiotemporal caching mechanism. Measurement results demonstrate end-to-end inference with 25.6$\mu$s latency and 1.7$\mu$J energy per event on state-of-the-art workloads such as DAGr-GNN, a 10-to-1000$\times$ improvement over prior art.%
\end{abstract}


%% file: Sections/intro.tex
\section{Introduction}
\begin{figure}[!t]
    \centering
    \includegraphics[width=\linewidth]{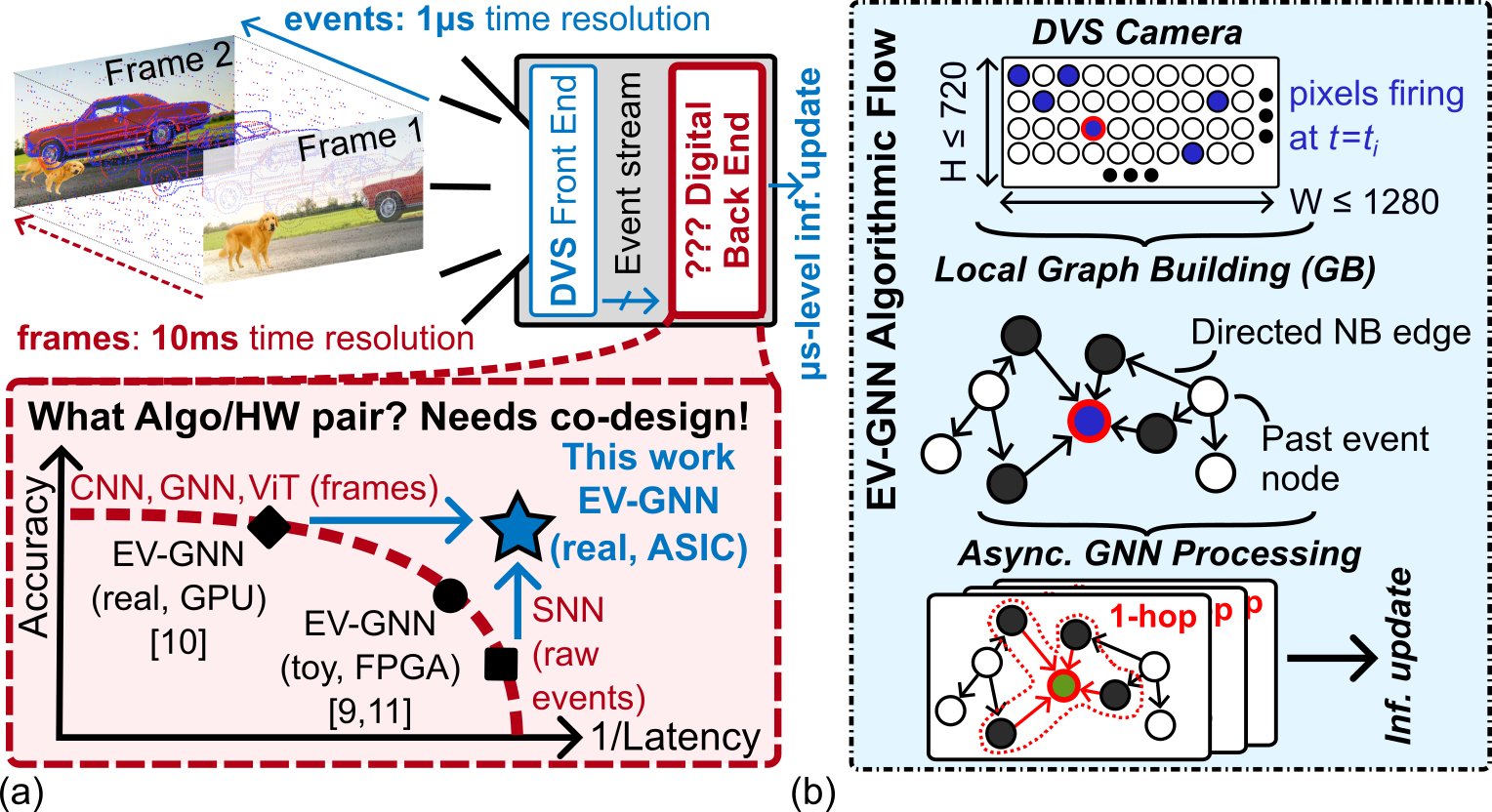}
    \caption{(a) Algorithm-hardware co-design challenge for state-of-the-art low-latency event-based processing. (b) Working principle of EV-GNNs.}
    \label{Fig_intro}
\end{figure}

\IEEEPARstart{U}{ltra-low-latency} detection is key for smart real-time edge applications, from safe autonomous car/drone navigation to virtual-reality headsets \cite{Gupta2021, Elbamby2018, Verma2024} (Fig. \ref{Fig_intro}(a)). Unlike conventional frame-based cameras, which have a latency of several ms, dynamic vision sensors (DVS) generate asynchronous event streams with a $\mu$s-level temporal resolution \cite{Lichtsteiner2008}. However, today’s vision processing systems fail to achieve both high accuracy and low latency \cite{Scherer2023, Viale2021, Frenkel2022}. Recently, event-driven graph neural networks (EV-GNNs) have emerged as a promising approach to bridge this gap by exploiting the sparsity and locality of event streams, first by adding each new event as a node of a spatiotemporal graph, then by using a GNN to exploit efficient local updates between the new node and its neighbors (NB) (Fig. \ref{Fig_intro}(b)) \cite{Yang2025, Gehrig2024}.
% State-of-the-art (SotA) EV-GNN detection networks, such as DAGr-GNN [10], combine 3D- and 2D-graph-convolutional layers (Fig. \ref{Fig_algo}). First, a few 3D layers with spatiotemporal information $(x, y, t)$ and a low number of channel features per graph node operate on the high-resolution graph-data maps, where neighborhoods are irregularly defined within a maximum spatiotemporal radius. Next, several 2D layers process lower-resolution voxel maps, where each voxel carries spatial-only information $(vx, vy)$ but contains many feature channels. In between, a 3D-to-2D graph-pooling layer handles the 2D projection into voxels, regularizing NB connections by limiting them to adjacent voxels.

%While a few EV-GNN accelerators have been proposed on FPGA [10-11], they are limited to low-resolution, toy setup that cannot be scaled up to such SotA networks due to three main challenges: (1) irregular external memory accesses (EMAs) to the graph’s 3D data maps lead to a sparse-irregular memory-bounded 3D regime, (2) inability to exploit the sparse-regular neighborhood of 2D maps fails to leverage high parallelism for low latency per inference under a compute-bounded 2D regime, and (3) no efficient HW support for spline-convolution operations impedes high-accuracy detection within footprints suitable for edge computing.

\begin{figure}[!t]
    \centering
    \includegraphics[width=\linewidth]{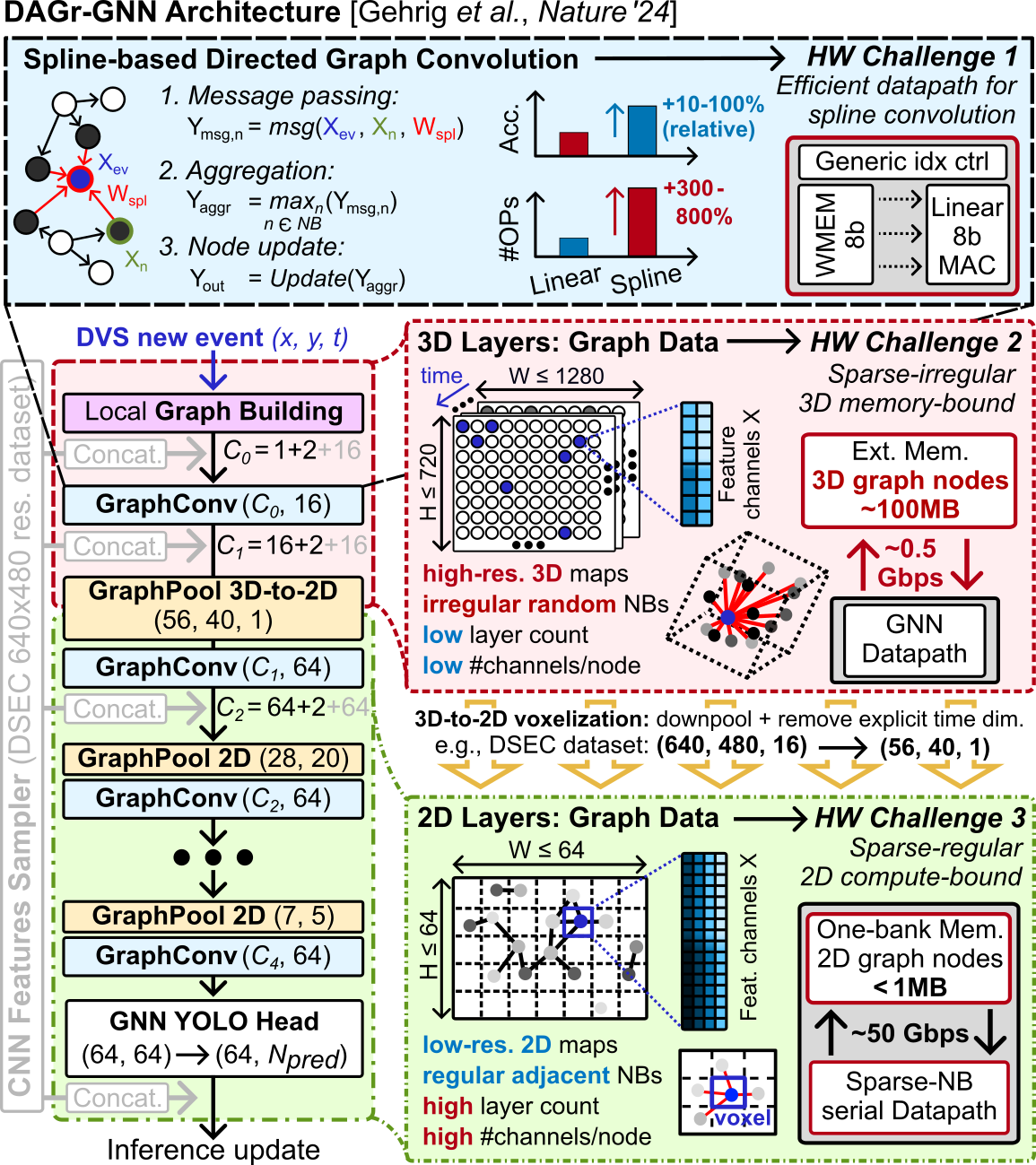}
    \caption{State-of-the-art DAGr-GNN workload and its HW challenges.}
    \label{Fig_algo}
\end{figure}

While a few EV-GNN accelerators have been proposed on FPGA \cite{Yang2025, Jeziorek2025}, they are limited to low-resolution toy setups, as three critical challenges impede hardware (HW) scalability to high-resolution workloads such as DAGr-GNN \cite{Gehrig2024} (Fig. \ref{Fig_algo}):\\
\textbf{(1)} Modern EV-GNNs rely on spline-convolution to encode positional information in the weights, which improves detection accuracy by up to 2$\times$. However, the lack of efficient HW support for spline-convolution operations, which involve up to 800\% more compute than linear graph convolution \cite{Matthias2018}, hinders their deployment within acceptable footprints.\\
\textbf{(2)} Such EV-GNNs first consist of a few 3D layers, which are memory-bound. Indeed, they rely on spatiotemporal $(x, y, t)$ high-resolution maps, with a low number of channel features per node. Neighborhoods are defined within a spatiotemporal radius, leading to significant external-memory accesses (EMAs) to the graph’s large-footprint 3D data maps that are both sparse and irregular.\\
\textbf{(3)} The 3D layers are followed by several 2D layers, after a pooling projection that removes the explicit time dimension. These 2D layers are compute-bound: they consist of lower-resolution \textit{voxel} maps, where each voxel $(vx, vy)$ carries spatial-only position information, but contains many feature channels as well as the overall graph connectivity. Although neighborhoods are limited to adjacent spatial voxels,% and thus regularly located in space,
their sparse connection is usually encoded as source-to-destination edge lists, which hinders leveraging high parallelism for low latency due to costly look-ups.

\begin{figure}[t!]
    \centering
    \includegraphics[width=\linewidth]{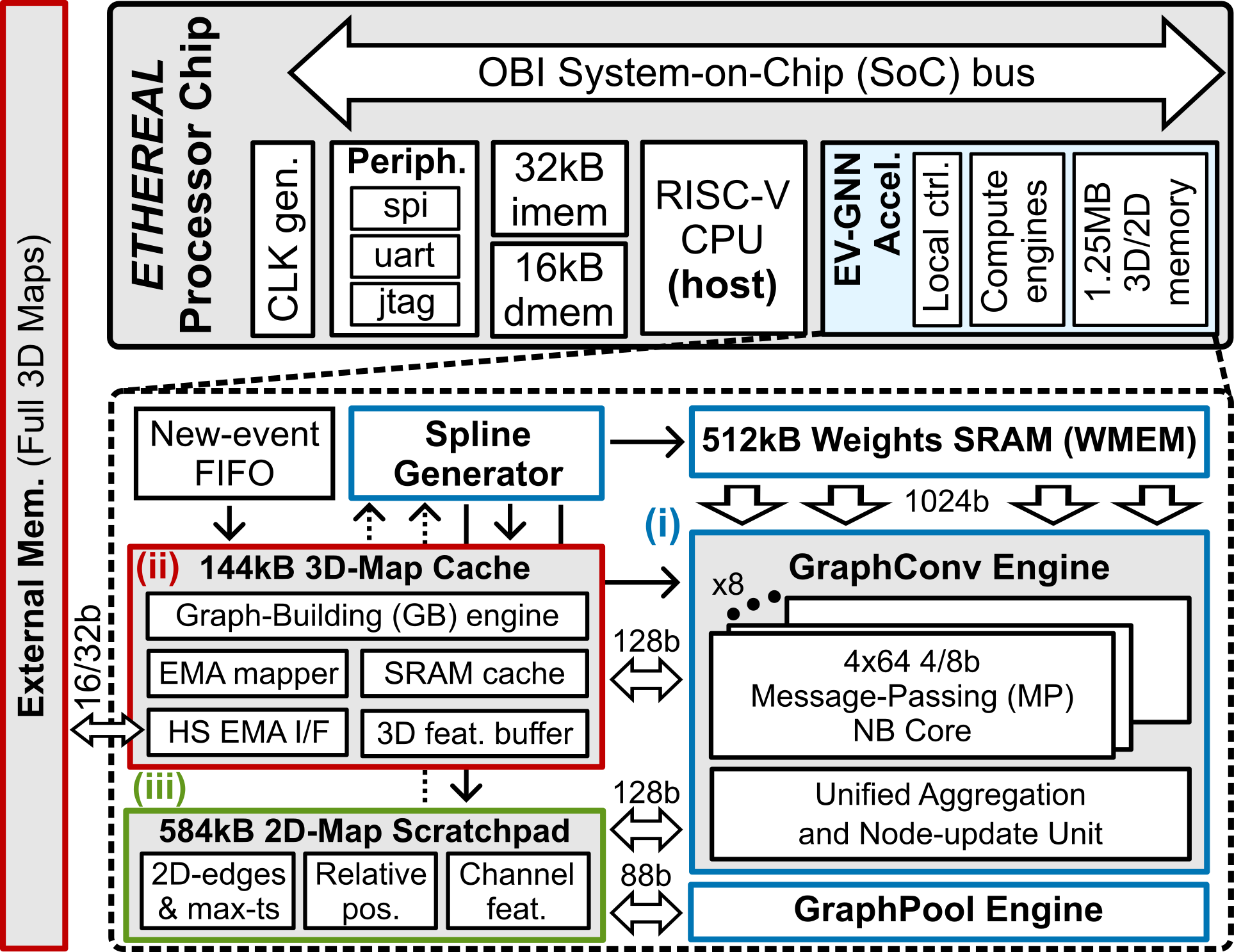}
    \caption{Overview of (a) the ETHEREAL chip and (b) its EV-GNN accelerator.}
    \label{Fig_arch_overview}
\end{figure}

To address these challenges, we propose ETHEREAL, the first EV-GNN processor chip (Fig. \ref{Fig_arch_overview}). Embedded in a low-footprint RISC-V-based SoC for control, its % loosely coupled
EV-GNN accelerator features (i) an output-parallel-and-stationary datapath with eight 4/8b-configurable spline-convolution message-passing (MP) cores followed by a unified aggregation and node-update unit, (ii) a 3D spatiotemporal data cache that leverages neighbors locality within regions of interest to reduce irregular EMAs, (iii) a fully on-chip 2D data scratchpad that enables NB-parallel processing through simultaneous multi-neighbor node-data access and one-hot encoding of the 2D-edge connectivity, thereby solving the three HW challenges above and unlocking scalability to modern EV-GNN workloads.

This paper is organized as follows: Section II presents ETHEREAL's event-driven spline convolution and its dedicated MP cores. Section III then details the 3D and 2D memory architectures, including memory-datapath dataflow scheduling. Finally, Section IV presents chip measurements.

%% file: Sections/datapath.tex
\section{Spline-convolution Dataflow and Datapath}

\begin{figure}[!t]
    \centering
    \includegraphics[width=\linewidth]{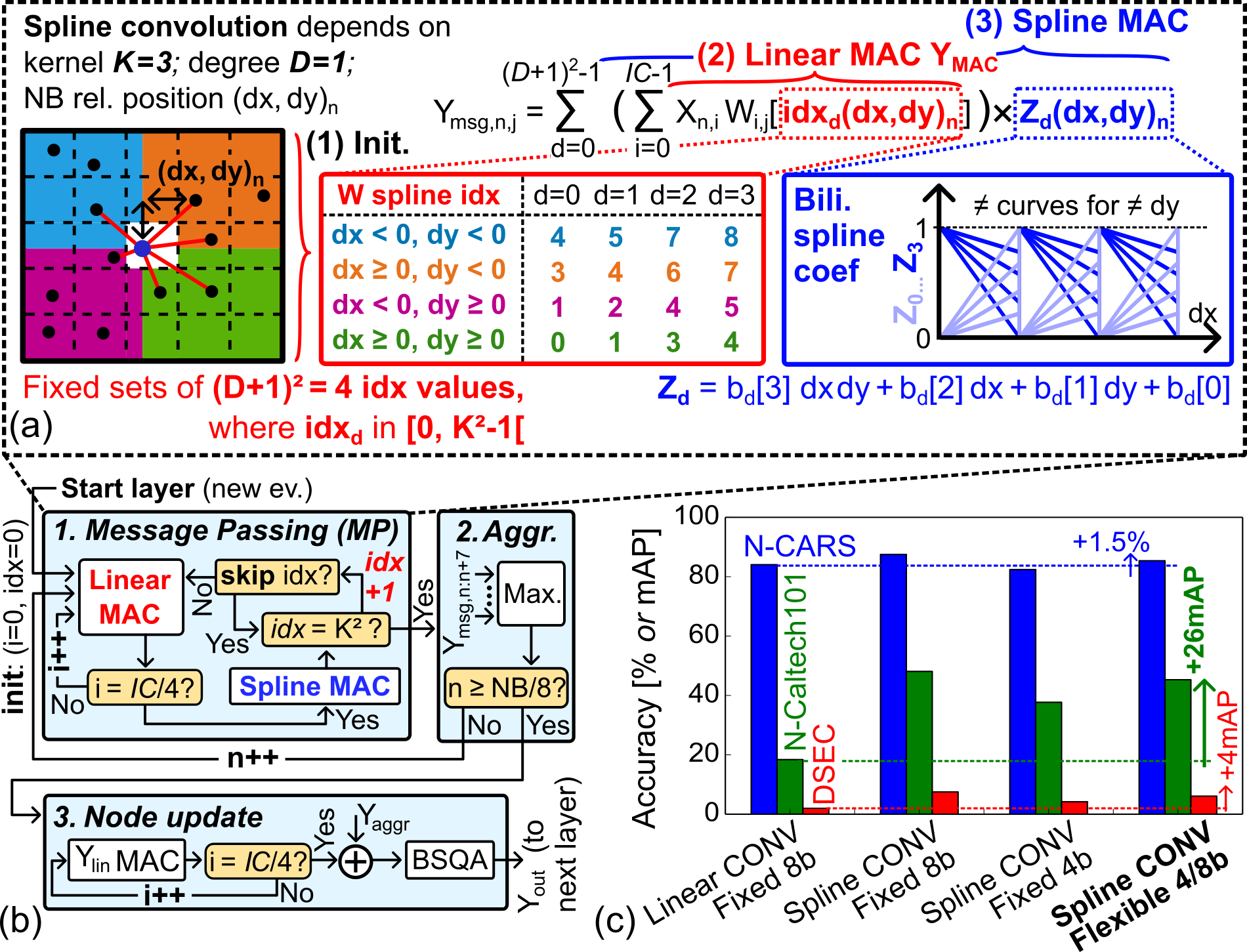}
    \caption{(a) Three-phase decomposition of spline-based MP into MACs, and (b) its iterative execution across $W$ spline indices in ETHEREAL's graph-convolution dataflow. (c) Accuracy benefit of precision-flexible splines.}
    \label{Fig_spline_conv}
\end{figure}

Conventional 2D/3D convolutions apply static filter weights to a pixel neighborhood, where all positions are predetermined.
In contrast, spline-based convolution embeds the positional difference $(dx, dy)$ between a node and its NB by modulating weights W with splines of a certain degree $D$ and kernel $K$ \cite{Matthias2018}, turning knowledge of positional information into accuracy gains. 
For 2D splines with $D = 1$ and $K = 3$, the MP output $Y_{msg}$ is a linear combination of $(D+1)^2 = 4$ base kernels. By expressing $Y_{msg}$ as a decomposition of regular MAC operations, we identify three phases (Fig.  \ref{Fig_spline_conv}(a)):
(1) an \textit{initialization} to find the position-dependent set of $(D+1)^2=4$ \textit{spline indices idx} that determine the weight tensors \textit{W[idx]} to apply to each NB; (2) a \textit{linear MAC} between the looked-up weights and input features \textit{X}, yielding an intermediate $Y_{MAC}$ result; (3) a \textit{spline MAC} between $Y_{MAC}$ and bilinear, position-dependent \textit{spline coefficients} $Z$ that apply a position-dependent modulation.
To leverage spatial weight reuse, we propose to directly iterate over the spline indices \textit{idx} %. to leverage spatial weight reuse %NB-level parallelism 
(Fig. \ref{Fig_spline_conv}(b)). Indeed, weights $W[idx]$ can be shared across different (active) MP cores at each iteration, % in parallel, 
thereby %avoiding repeated accesses to the same $W$ value and 
maximizing utilization of the 1024b-bandwidth WMEM. This \textit{spline-iterative} MP phase is followed by the aggregation and node update phases, yielding the spline-convolution output features $Y_{out}$ in Fig. \ref{Fig_spline_conv}(b).

\begin{figure}[!t]
    \centering
    \includegraphics[width=\linewidth]{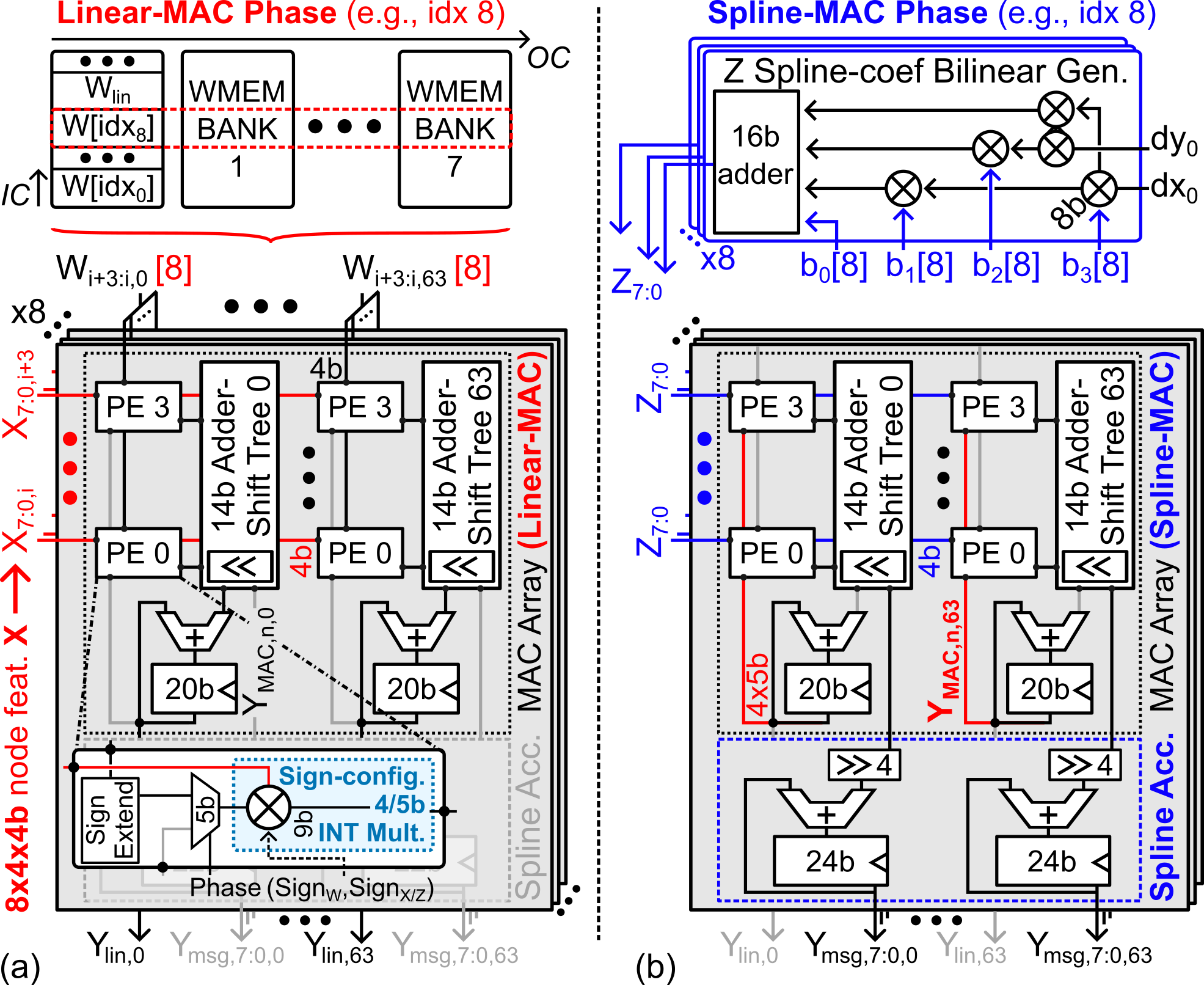}
    \caption{Reconfigurable MP cores during (a) linear and (b) spline MAC phases.}
    \label{Fig_datapath}
\end{figure}

We exploit this dataflow in both the 3D and 2D layers in ETHEREAL's reconfigurable graph-convolution %engine’s 
datapath, which features eight NB-parallel MP cores (Fig. \ref{Fig_datapath}(a)-(b)) whose outputs gather into a unified aggregation and node-update unit. % (BSQA).
Each MP core consists of a 4$\times$64 configurable-MAC array with 5/4b processing elements (PEs) followed by spline accumulators that add four $Y_{MAC}$ values into an output message $Y_{msg}$. Each PE embeds a linear multiplier whose sign and operands change with the MP phase. During the linear-MAC phase (Fig. \ref{Fig_datapath}(a)), signed 4b weights \textit{W} are broadcast to all MP cores in parallel, while 8$\times$4 4b (un)signed NB inputs \textit{X} are fed to the PEs. In this phase, mixed-4/8b operations are supported by multi-cycle MACs with post-accumulation left-shift bit alignment: this configurability is key to preserve high accuracy with minimum latency and memory footprints (Fig. \ref{Fig_spline_conv}(c)). During the spline-MAC phase (Fig. \ref{Fig_datapath}(b)), the signed 20b partial sum $Y_{MAC}$ is split column-wise across PEs in four packs of 5b, while the second input operand of the PEs is now fed with the unsigned spline coefficients \textit{Z}. These are generated by a bilinear-interpolation unit based on the relative position \textit{(dx, dy)} of NB nodes, and precompiled parameters \textit{b[idx]}. The two-mode configurability limits the MP core's area overhead compared to a linear MAC array below 30\%. 

%Finally, the unified BSQA units generate the sum-of-term outputs $Y_{out}$ expected after the aggregation and node-update phase (Fig. \ref{Fig_datapath}(c)). Importantly, an error-correction circuit keeps track of the total residue of the three-phase mixed INT/FP-to-INT rescaling ($S_1, S_2, S_3$), and compensates for it during the last BSQA phase, before quantization. This mechanism prevents output-error accumulation over time, which is critical in the state-based EV-GNNs.

%% file: Sections/mem.tex
\section{Split-3D/2D Data Memory System}
To address 3D sparse-irregular and 2D sparse-regular node-data accesses, we split the EV-GNN memory in two parts.

\begin{figure}[!t]
    \centering
    \includegraphics[width=\linewidth]{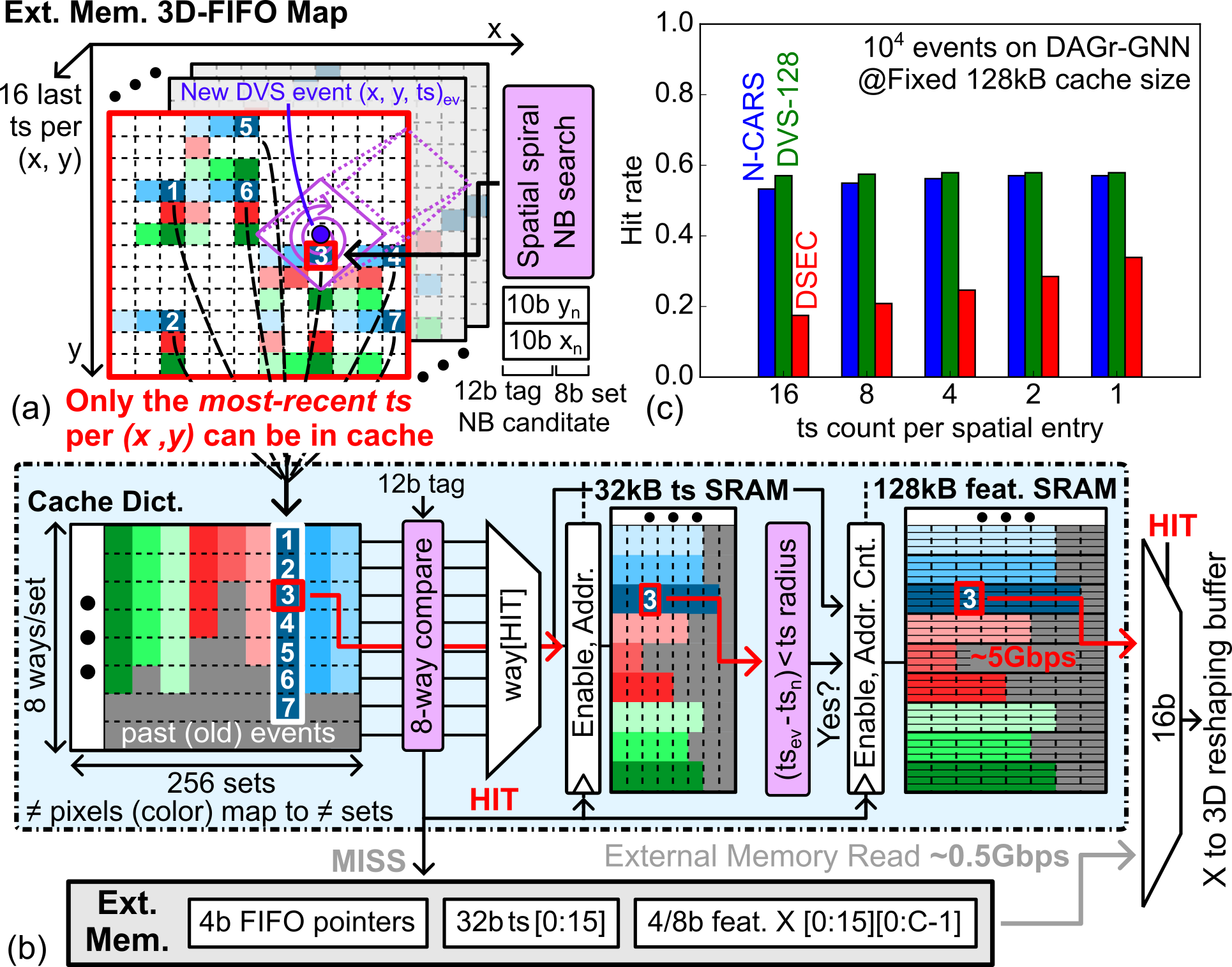}
    \caption{(a) Illustration of spatiotemporal locality (RoIs) in the external memory's 3D-FIFO maps, and example of spatial NB search for a new event. (b) 3D cache architecture, illustrating the most-recent-timestamp ($ts$), 8-spatial-ways mapping policy, with a three-phase hit-read scheme that intertwines temporal search. (c) Hit rate for different datasets and caching schemes.}
    \label{Fig_cache}
\end{figure}

\begin{figure}[!t]
    \centering
    \includegraphics[width=\linewidth]{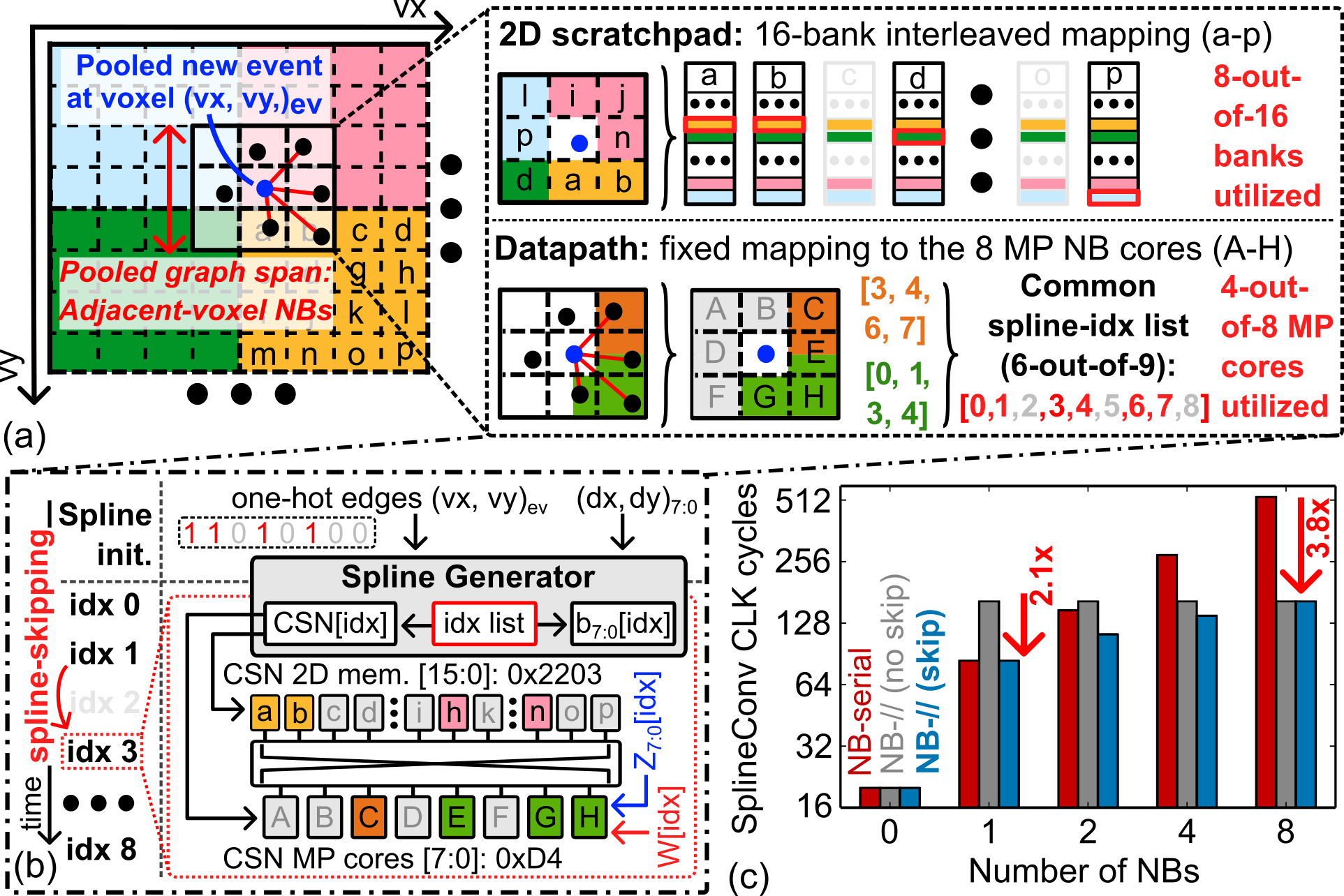}
    \caption{(a) Mapping of a new 2D voxel event and its neighbors to 2D memory and datapath cores. (b) Illustration of the corresponding spline-iterative dataflow with skipping. (c) Latency benefits of spline skipping.}
    \label{Fig_spad}
\end{figure}

\textbf{3D memory:} we leverage the spatiotemporal locality of NBs in the event graph to reduce EMAs in the memory-bounded 3D layers. 
To that end, we introduce an 8-way set-associative 3D cache (256 sets/way), %8-spatial ways, 256-set-associative 3D cache 
that keeps track of event data from regions of interest (RoIs) in the event-stream graph (Fig. \ref{Fig_cache}(a)).
%This topology was motivated by finding a balance between cache hit rate and the area/energy overhead of the cache lookup process.
The 3D-cache architecture is intertwined with the EV-GNN's graph-building process, similar to \cite{Yang2025} (Fig. \ref{Fig_cache}(a)-(b)). 
Upon the reception of a new event at location $(x, y)$, a new spatial candidate is derived from a spiral-like spatial NB search. First, a cache lookup compares the NB candidate's spatial tag with that of all ways of the indexed set. % of the target dictionary set. 
Then, upon a hit, a temporal search compares the cache entry's timestamp ($ts$) with that of the new event: if their difference lies within a user-defined temporal radius, the NB candidate is deemed valid. Its features $X$ are then finally transferred at 5Gbps to a 3D buffer for NB-parallel reshaping, before being streamed to the graph-convolution engine. Upon a spatial miss, the search process triggers EMAs at a 0.5Gbps bandwidth, thereby making the NB search 10$\times$ slower.
This bandwidth gap underlines the importance of maximizing cache hits to minimize the latency overhead. Analyzing edge-vision datasets with various DVS camera resolutions, we found out that storing only the most-recent $ts$ per $(x, y)$ location usually maximizes the hit rate, at a fixed cache size and number of ways (Fig. \ref{Fig_cache}(c)). Hit rates reaching up to 58\% turn into a $2.4\times$ reduction in read EMAs, thereby significantly alleviating the memory-bound regime of 3D layers.

\textbf{2D memory:} we leverage the regularity of spatial-only neighborhoods in the 2D layers to foster high parallelism and avoid EMAs altogether. To that end, we take advantage of the observed spatial-adjacency property of NB candidates in 2D voxel maps: node data are mapped to a 16-bank memory in a 2D-interleaved manner, whereas processing of each of the eight potential NBs is attributed to a fixed MP core (Fig. \ref{Fig_spad}(a)). This mapping enables peak parallelism across neighbors by continuously streaming 8$\times$16b 2D node features \textit{X} per cycle from the 2D memory.
Moreover, by directly accessing the relative NB-to-event position $(dx, dy)$ and the set of inbound 2D neighbor edges in the 2D memory, the spline generator initializes the list of valid spline indices in a single cycle, as opposed to the 3D case that necessitates prior graph building. Additional latency and energy gains are respectively leveraged by skipping unnecessary spline indices during the spline-iterative MP phase, and by clock gating unused memory banks/MP cores (Fig. \ref{Fig_spad}(b)). This dataflow improves latency per inference by up-to 3.8$\times$ compared to a NB-serial spline-convolution (extrapolated from \cite{Yang2025}) (Fig. \ref{Fig_spad}(c)). 

%% file: Sections/meas.tex
\section{Measurement Results}
\begin{figure}[!t]
    \centering
    \includegraphics[width=\linewidth]{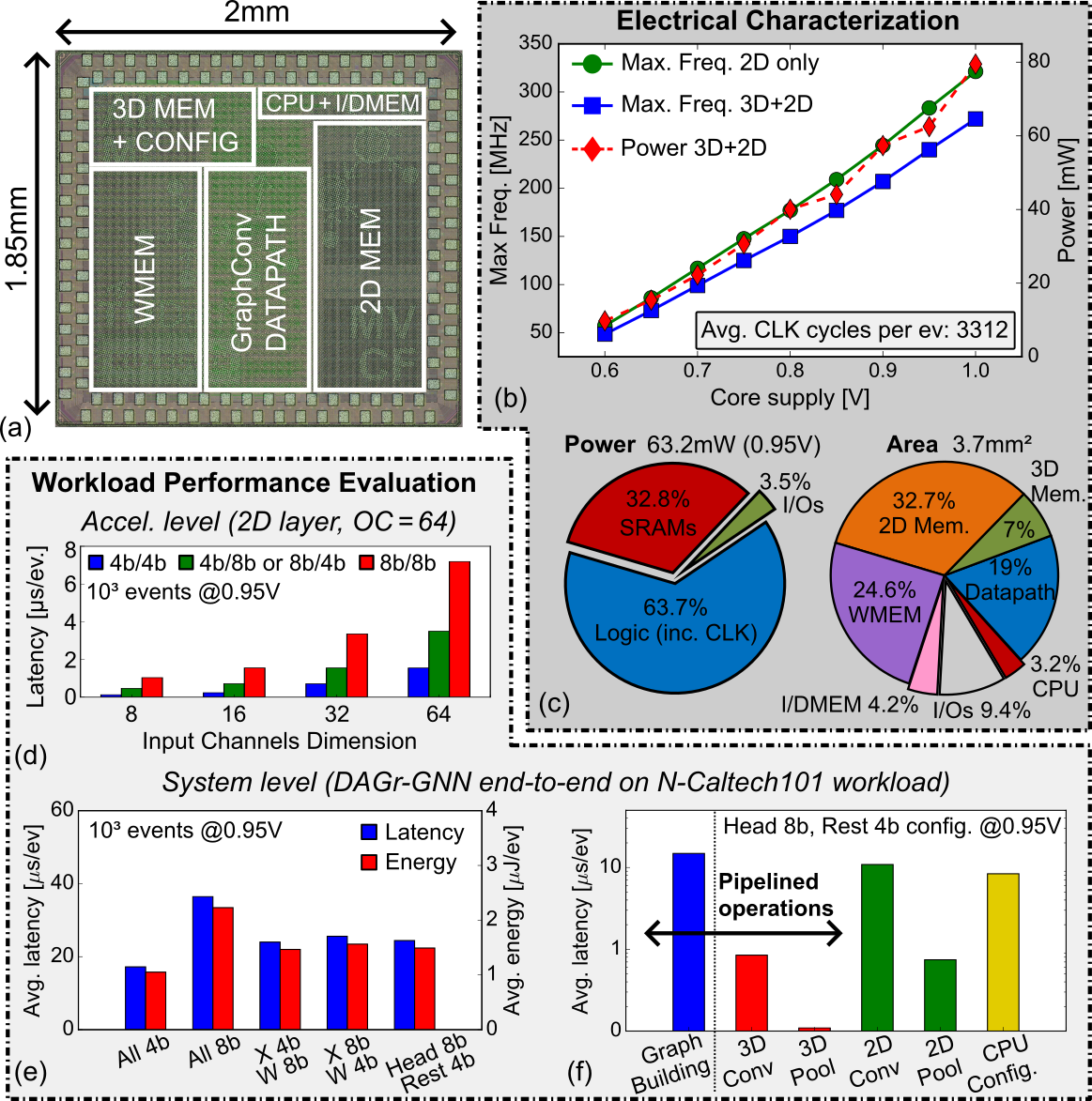}
    \caption{Measurements: (a) chip microphotograph, (b)-(c) electrical performance, and workload evaluation at the accelerator (d) and system (e)-(f) levels.}
    \label{Fig_meas}
\end{figure}

The ETHEREAL chip has been implemented in a TSMC 28nm process (Fig. \ref{Fig_meas}(a)) and reaches up to 250MHz at a target 0.95V on DAGr-GNN (Fig. \ref{Fig_meas}(b)). ETHEREAL's total power of 63.2mW is then dominated by logic operations, whereas SRAMs occupy the majority of the chip area (Fig. \ref{Fig_meas}(c)).

Performance evaluation of ETHEREAL on the DAGr-GNN DVS workload demonstrates its flexible trade-off between latency, energy and accuracy. 
At the accelerator level (Fig. \ref{Fig_meas}(d)), the per-layer latency ranges from 0.3 to 7.1$\mu$s for a representative 64$\times$48 voxel map, showcasing performance adaptivity to various channel dimensions and bit precision.
At the system level (Fig. \ref{Fig_meas}(e)), the layer-wise configuration of the bit precision determines the end-to-end performance (related to the accuracy target in Fig. \ref{Fig_spline_conv}(c)), yielding a per-event latency (resp. energy) between 17.8 and 36.5$\mu$s (resp. 1.1 and 2.3$\mu$J) at 0.95V. An end-to-end latency breakdown (Fig. \ref{Fig_meas}(f))) highlights the balance between processing and configuration arising from the event-driven dataflow, where graph building is pipelined with other operations to reduce the total latency by 1.8$\times$.

\input{Tabs/SoA}

Compared to the SotA (Table \ref{Tab1}), ETHEREAL is the only design to achieve a $\mu$s-level inference on deep networks, improving the detection latency by up to 1000$\times$ over prior works while achieving comparable accuracy. These include existing EV-GNN accelerators evaluated on N-CARS and N-Caltech101, which are limited to low-resolution ($\leq$240$\times$180) DVSes. In contrast, ETHEREAL is the first work to demonstrate scalability toward high-resolution DVS by mapping DSEC (640$\times$480) with records 25.6$\mu$s latency and 1.7$\mu$J energy per event inference.

%% file: Tabs/SoA.tex
\begin{table}[!t]
\centering
\renewcommand{\arraystretch}{1.1}
\setlength{\tabcolsep}{3pt}

\caption{Comparison to the state of the art}
\label{Tab1}

\resizebox{0.5\textwidth}{!}{%

% Column spec:
% First column left-aligned with wrapping
% All other columns centered with wrapping
\begin{tabular}{
    >{\raggedright\arraybackslash}p{2.0cm}|   % descriptor column
    >{\centering\arraybackslash}p{1.2cm}|    % GPU
    >{\centering\arraybackslash}p{1.0cm}    % SNN 1
    >{\centering\arraybackslash}p{0.9cm}    % SNN 2
    >{\centering\arraybackslash}p{1.0cm}|   % SNN 3
    >{\centering\arraybackslash}p{1.0cm}    % EV-GNN 1
    >{\centering\arraybackslash}p{1.2cm}    % EV-GNN 2
    >{\centering\arraybackslash}p{2.2cm}    % EV-GNN 3 (This work)
}
\toprule
            & \multicolumn{1}{|c|}{\textbf{GPU}} 
            & \multicolumn{3}{c|}{\textbf{SNN}}
            & \multicolumn{3}{c}{\textbf{EV-GNN}} 
            \\ \midrule
            & Measured & {\cite{Viale2021}\textsuperscript{(iii)}} 
            & {\cite{Frenkel2022}\textsuperscript{(iii)}} 
            & {\cite{Fang2025}\textsuperscript{(iii)}} 
            & {\cite{Yang2025}} 
            & {\cite{Jeziorek2025}} & \textcolor{blue}{\textbf{This Work}\textsuperscript{(iii)}}  \\
\midrule

Year               & 2026 & 2021 & 2022 & 2024 & 2025 & 2025 & 2026 \\
Technology         & A100 & 14nm & 28nm & 40nm & FPGA & FPGA & 28nm \\
Supply [V]         & 0.9 & 0.8  & 0.5-0.8  & --   & 0.8  & 0.8  & 0.6-1 \\
Freq. [MHz]        & -- & 60 & 13-115 & 50-200 & 200 & NA & 280 \\
Total Memory       & -- & 27MB & 138kB & -- & 780kB & NA & 1.25MB \\
Area [mm$^2$]      & -- & 60 & 0.9 & 2.2 & -- & -- & 3.7 \\
X/W Precision      & FP64 & 1/8b & 1/8b & 1-8/8b & 8b & 8b & 4-8/4-8b \\
Max. Img. Res.     & 640$\times$480 & 128$\times$128 & 128$\times$128 & 128$\times$128 & 120$\times$100 & 240$\times$180 & \textbf{\color{blue}{640$\times$480}} \\
\midrule

\, \newline \textit{Repr. workload} \newline \newline Acc. on HW %\newline (SW baseline
&
\textit{N-CARS} \newline 90.7\textsuperscript{(iv)}  
\textit{N-Cal.101} \newline 52.6\textsuperscript{(iv)}  
\textit{DSEC} \newline 12.4\textsuperscript{(iv)}
& 
\textit{N-CARS} \newline 94.5 \newline \textit{DVS128} \newline 90.5 \newline -- \newline \,
&
-- \newline \, \newline \textit{DVS128}\newline 87.3  \newline -- \newline \,
&
-- \newline \, \newline \textit{DVS128}\newline 96.1  \newline -- \newline \,
&
\textit{N-CARS} \newline 87.8 \newline -- \newline \, \newline -- 
&
\textit{N-CARS} \newline 92.5\newline
\textit{N-Cal.101} \newline 62.8 \newline
-- \newline \,
&
\textit{N-CARS} \newline 86.7\textsuperscript{(iv),(v),(vi)}\newline
\textit{N-Cal.101} \newline 48.1\textsuperscript{(iv),(v),(vi)}\newline
\textit{DSEC} \newline 7.6\textsuperscript{(iv),(v),(vi)}
\\
\midrule

Peak Thrput.\newline [TOPS/b]\textsuperscript{(i),(ii)}
& -- & 0.04 & 0.04 & 92 & -- & 0.24 & 16 \\

Peak Ene. Eff. \newline [TOPS/W/b]\textsuperscript{(i),(ii)}
& -- & 0.48 & 12 & 2740 & -- & -- & 113 \\

Avg. Latency/inf.
& 140ms
& 900$\mu$s 
& 600$\mu$s 
& 2.2ms 
& 16$\mu$s
& $\geq$ 5.1ms 
& \textbf{\textcolor{blue}{17-36$\mu$s\textsuperscript{(iv),(v)}}} \\

Avg. Energy/inf.
& 400mJ & 320$\mu$J & 46$\mu$J & 6.2$\mu$J & -- & -- & \textbf{\textcolor{blue}{1.1-2.3$\mu$J\textsuperscript{(iv),(v)}}} \\
\bottomrule
\end{tabular}

}

%\vspace{0.1cm}

\flushleft

\scriptsize{
(i) Linearly normalized to 1b inputs and weights.\,
(ii) 1 MAC = 2 OP.\,
(iii) System level (excl.\ DRAM).\,
(iv) DAGr-S, GNN only (directed) \,
(v)  Mixed 4/8b with QAT (0.95V).
(vi) End-to-end acc.\ simulated with HW model, measured end-to-end validity on synth. data.
}

\end{table}

%% file: Sections/conclusion.tex
\section{Conclusion}
In this work, we presented ETHEREAL, the first EV-GNN processor for ultra-low-latency, high-resolution edge vision. To address the three scalability challenges of modern EV-GNNs, it features a precision-flexible spline-convolution datapath, a 3D spatiotemporal cache for regions of interest, and a 2D-interleaved memory that enables NB parallelism. Measurement results showcase a 25.6$\mu$s latency and a 1.7$\mu$J energy per inference on the SotA DAGr-GNN for DSEC. ETHEREAL thereby achieves a 10-to-1000$\times$ improvement over prior designs, and is the first to scale up to 640$\times$480 DVS inputs. 